\documentclass[runningheads]{llncs}
\usepackage{graphicx}      
\usepackage{amsmath,amssymb,amsfonts}
\usepackage{bm, algorithmic}
\usepackage{textcomp}
\usepackage{xcolor}

\usepackage[ruled]{algorithm2e}
\usepackage{nicefrac}

\begin{document}

\title{On Solving a Stochastic Shortest-Path Markov Decision Process as Probabilistic Inference}

\titlerunning{SSP MDP as Probabilistic Inference}

\author{Mohamed Baioumy \and Bruno Lacerda\and Paul Duckworth \and Nick Hawes}


\authorrunning{M. Baioumy et al.}

%


\institute{Oxford Robotics Institute, University of Oxford\\
	\email{\{mohamed, bruno, pduckworth, nickh\}@robots.ox.ac.uk}}


\maketitle              
%
\begin{abstract}
Previous work on planning as active inference addresses finite horizon problems and solutions valid for \textit{online} planning.  We propose solving the general Stochastic Shortest-Path Markov Decision Process (SSP MDP) as probabilistic inference. Furthermore, we discuss online and offline methods for planning under uncertainty. In an SSP MDP, the horizon is \textit{indefinite} and unknown a priori. SSP MDPs generalize finite and infinite horizon MDPs and are widely used in the artificial intelligence community. Additionally, we highlight some of the differences between solving an MDP using dynamic programming approaches widely used in the artificial intelligence community and approaches used in the active inference community. F
\end{abstract}

\section{Introduction}

A core problem in the field of artificial intelligence (AI) is building agents capable of automated planning under uncertainty. Problems involving planning under uncertainty are typically formulated as an instance of a Markov Decision Process (MDP). At a high level, an MDP comprises 1) a set of world states, 2) a set of actions, 3) a transition model describing the probability of transitioning to a new state when taking an action in the current state, and 4) an objective function (e.g. minimizing costs over a sequence of time steps). An MDP solution determines the agent’s actions at each decision point. An optimal MDP solution is one that optimizes the objective function. These are typically obtained using dynamic programming algorithms \footnote{Linear programming approaches are also popular methods for solving MDPs \cite{bertsekas1991sss_analysis,forejt2011automated,nazareth1986linear_programming,d1963probabilistic}. Additionally, other methods exist in the reinforcement learning community such as policy gradient methods \cite{thomas2017policy,grondman2012survey,sutton2000policy}. } \cite{kolobov2012planning,sutton1998introduction}. 

Recent work based on the active inference framework \cite{friston2017active} poses the planning problem as a probabilistic inference problem. Several papers have been published showing connections between active inference and dynamic programming to solve an MDP \cite{kaplan2018planning_AIF,da2020relationship,da2020active}. However, the planning problem being solved in the two communities is not equivalent. 


First, dynamic programming approaches used to solve an MDP, such as policy iteration, are valid for finite, infinite and indefinite horizons. Indefinite horizons are finite but of which the length is unknown a priori. For instance, consider an agent navigating from a starting state to a goal state in a grid world where the outcome is uncertain (e.g. the $4\times4$ grid world in Figure \ref{fig:plan_vs_policy} shown in the appendix). Before starting to act in the environment, there is no way for the agent to know how many time steps it will take to reach the goal. Algorithms based on dynamic programming, such as policy iteration, are valid for such settings. They can solve the Stochastic Shortest-Path Markov decision process (SSP MDP)\cite{kolobov2012planning,bertsekas1991sss_analysis}. However, work from active inference is only formulated for finite horizons \cite{da2020relationship}.

Second, the optimal solution to an SSP MDP is a stationary deterministic policy \cite{kolobov2012planning}. This refers to a mapping from states to actions independent of time. Computing this optimal policy can be done \textit{offline} (without interaction with the environment) or online (while interacting). In the active inference literature however, solving the planning problem is performed by computing a stochastic plan (a sequence of actions given the current state). This is only valid during online planning. Additionally, the solution is only optimal given a certain horizon, which is specified a priori. If the horizon chosen is too short, the agent will not find a solution to reach the target. If it is too long, the solution will be sub-optimal. 



The main contribution of this paper is presenting a novel algorithm to solve a Stochastic Shortest-Path Markov Decision Process using probabilistic inference. This is an MDP with an \textit{indefinite horizon}. Additionally, highlighting the several gaps between solving an MDP in the AI community and the active inference community. 

Section \ref{sec:ssp_and_dynamic_programming} discusses the SSP MDP and section \ref{sec:ssp_as_inference} presets an approach for solving an SSP MDP as probabilistic inference. The equivalence between the two methods is shown in Section \ref{subsec:connections_btween_the_2_views}. Furthermore, the difference between world states and temporal state is highlighted in Section \ref{subsec: temporal_vs_world_states}. Policies, plans and probabilistic plans are discussed in Section \ref{subsec:plan_policy_prob_plan}.  Finally, a discussion on online vs offline planning can be found in Section \ref{sec:discussion}.

\section{Stochastic Shortest Path MDP}
\label{sec:ssp_and_dynamic_programming}

An SSP MDP is defined as a tuple $\mathcal{M}=$ $\left(S, A, C, T, G\right)$. $S$ is the set of states, $A$ is the set of actions, $C: S \times A \times S \rightarrow \mathbb{R}$ is the cost function, and $T: S \times A \times S \rightarrow[0,1]$ is the transition function. $G \subset S$ is the set of goal states. Each goal state $s_{g} \in G$ is absorbing and incurs zero cost. The expected cost of applying action $a$ in state $s$ is $\bar{C}(s, a)=\sum_{s^{\prime} \in S} T\left(s, a, s^{\prime}\right) \cdot C\left(s, a, s^{\prime}\right) .$ The minimum expected cost at state $s$ is $\bar{C}^{*}(s)=\min _{a \in A} \bar{C}(s, a)$. A policy maps a state to a distribution over action choices. A policy is deterministic if it chooses a single action at each step. A policy $\pi$ is proper if it reaches $s_{g} \in G$ starting from any $s$ with probability 1. In an SSP MDP, the following assumptions are made \cite{kolobov2012planning}: a) there exists a proper policy, and b) every improper policy incurs infinite cost at all states where it is improper.

The goal is to find the an \textit{optimal policy} $\pi^*$ with the minimum expected cost $\bar{C}^{*}(s)$ and can be computed as

$$
\pi^{*}(s)=\operatorname{argmin}_{a \in \mathcal{A}}\left[\sum_{s^{\prime} \in \mathcal{S}} \mathcal{T}\left(s, a, s^{\prime}\right)\left[\mathcal{C}\left(s, a, s^{\prime}\right)+V^{*}\left(s^{\prime}\right)\right]\right] .
$$

\noindent $V^{*}(s)$ is referred to as the optimal value for a state $s$ and is defined as:

$$
V^{*}(s)=\min _{a \in \mathcal{A}}\left[\sum_{s^{\prime} \in \mathcal{S}} \mathcal{T}\left(s, a, s^{\prime}\right)\left[\mathcal{C}\left(s, a, s^{\prime}\right)+V^{*}\left(s^{\prime}\right)\right]\right].$$

Crucially, the optimal policy $\pi^{*}$ corresponding to the optimal value function is Markovian (only dependant on the current state) and deterministic \cite{kolobov2012planning}. Solving an SSP MDP means finding a policy that minimizes expected cost, as opposed to one that maximizes reward. This difference is purely semantic as the problems are dual. We can define a reward function $\mathcal{R} = -\mathcal{C}$ and move to a reward maximization formulation. A more fundamental distinction is the presence of a special set of (terminal) goal states, in which staying forever incurs no cost.

Solving an SSP MDP can be done using standard dynamic programming algorithms such as policy iteration. Policy iteration can be divided in two steps, policy evaluation and improvement. In policy evaluation, for a policy $\pi$, the value function $V_{\pi}(s)$ is recursively evaluated until convergence as 
$$
\begin{aligned}
V_{\pi}(s) \leftarrow \sum_{s^{\prime} \in \mathcal{S}} \mathcal{T}\left(s, \pi(s), s^{\prime}\right)\left[\mathcal{C}\left(s, \pi(s), s^{\prime}\right)+V_{\pi}\left(s^{\prime}\right)\right]. 
\end{aligned}
$$

In the policy improvement step, the state-action value function $Q(s, a)$ is computed as: $$Q(s, a) = \sum_{s^{\prime} \in \mathcal{S}} \mathcal{T}(s, a, s^{\prime})[\mathcal{C}(s, a, s^{\prime})+V(s^{\prime})].$$ 

Then we compute a new policy as $\pi' = \operatorname{argmin}_{a \in \mathcal{A}}Q(s, a)$ for every state in $S$. Iterating between these two steps guarantees convergence to an optimal policy.

\subsubsection{Properties of an SSP MDP.} An SSP MDP can be shown to generalize finite, infinite and indefinite horizon MDPs \cite{bertsekas1995neuro,kolobov2012planning}. Thus algorithms valid for an SSP MDP are also valid for the finite and infinite horizon MDPs. Additionally, it can be proven that each SSP MDP has an optimal deterministic policy independent of time. Therefore, the claims made in \cite{da2020relationship} about active inference being more general since it computes stochastic policies are unjustified when solving an MDP. However, these results do not hold in partially observable cases or in the presence of uncertain models. But in the SSP MDP defined above (which is commonly used in the AI community), there always exists a deterministic optimal policy. Note that there is an infinite number of stochastic policies but only a finite number of deterministic policies ($|\mathcal{S}|^{\mathcal{|A|}}$). This greatly speeds up the algorithms while still guaranteeing optimality. 

\section{Solving an SSP MDP as probabilistic inference}
\label{sec:ssp_as_inference}

In this section we discuss a novel approach for solving an SSP MDP as probabilistic inference. We use an inference algorithm that exactly solves an SSP MDP as defined in the previous section. This approach is inspired by work 
from \cite{toussaint2006probabilistic,toussaint2006probabilisticPOMDP} which solves an MDP with an indefinite horizon. This approach has been successfully applied to solve problems of planning under uncertainty, e.g. \cite{kumar2015probabilistic,toussaint2008hierarchical}.

\subsection{Definitions}
\label{subsec:ssp_inference_definitions}
The definition of an SSP MDP includes a set of world states $\mathcal{S}$ and actions $\mathcal{A}$. In probabilistic inference we instead reason about \textit{temporal} states and actions. A temporal state $s_t$ is a random variable defined over all world states. Conceptually it represents the state that the agent will visit at the time-step $t$. For the grid world in Figure \ref{fig:plan_vs_policy}, there are 16 world states but the number of temporal states is unknown a priori since the horizon is unknown.

The transition probability is defined as a probability distribution over temporal states and actions as $P(s_{t+1}|a_t, s_t)$. If the random variables are fixed to specific world states $i$ and $j$ and an action $a$, the transition probability $P(s_{t+1}=j|a_t=a, s_t=i)$ would be equivalent to the transition function $\mathcal{T}$ defined for an SSP MDP. The probability of taking a certain action in a state is parameterized by a policy $\pi$ as $P(a_t=a|s_t=i;\pi) = \pi_{ai}$. This policy is defined exactly the same as in the case of an SSP MDP.

The cost function $P(c_t|s_t, a_t)$ is defined differently. The temporal cost variables $c_t$ are defined as binary random variables $c_t \in \{0, 1\}$. Translating an arbitrary cost function to temporal costs can be done by scaling the cost function $\mathcal{C}(s, a)$ (as defined in the previous section) between the minimum cost (min($\mathcal{C}$)) and maximum cost (max($\mathcal{C}$)) as: 

$$
P(c_t = 1 \mid a_t = a, s_t = s)=\frac{\mathcal{C}(a, s)-\min (\mathcal{C})}{\max (\mathcal{C})-\min (\mathcal{C})}.$$

Any expression with $P(c_t = 1)$ can be though of as `the probability of a cost being maximal'. Thus, the probability of a cost being maximal given a state and an action is $P(c_t = 1 \mid a_t = a, s_t = s)$.  Now we can reason about the highest possible cost for a state $s$ and action $a$ as one where $P(c_t = 1 \mid a_t = a, s_t = s) = 1$ and the lowest possible cost as $P(c_t = 1\mid a_t = a, s_t = s)$ = 0. Any other cost will have a probability in-between, according to its magnitude.

Finally, we model the horizon as a random variable. The temporal states and actions are considered up to the end of the horizon $T$. However, the horizon is generally unknown. We thus model $T$ itself as a random variable. Combining all this information we can define the SSP MDP using a probabilistic model.


\subsection{Mixture of finite MDPs}
In this section we define the SSP MDP in terms of a mixture of finite MDPs with only a final cost variable. Given every horizon (for instance $T = 1$) the finite MDP can be given as $P\left(c, s_{0: T}, a_{0: T} \mid T ; \boldsymbol{\pi}\right)$. Note that we dropped the time-index for $c_t$ since there is only one cost variable now.  This model can be factorized as

$$
\begin{array}{r}
P\left(c, s_{0: T}, a_{0: T} \mid T ; \boldsymbol{\pi}\right)= P\left(c \mid a_{T}, s_{T}\right) P\left(a_{0} \mid s_{0} ; \boldsymbol{\pi}\right) \\ P\left(s_{0}\right) 
\cdot \prod_{t=1}^{T} P\left(a_{t} \mid s_{t} ; \boldsymbol{\pi}\right) P\left(s_{t} \mid a_{t-1}, s_{t-1}\right)
\end{array}
$$

To reason about the full MDP, we consider the mixture model of the joint given by the joint probability distribution
$$
P\left(c, s_{0: T}, a_{0: T}, T ; \boldsymbol{\pi}\right)=P\left(c, s_{0: T}, a_{0: T} \mid T ; \boldsymbol{\pi}\right) P(T)
$$
where $P(T)$ is a prior over the total time, which we choose to be a flat prior (uniform distribution).



\subsection{Computing an optimal policy}
Our objective is to find a policy that minimizes the expected cost. Similarly to policy iteration, we do not assume any knowledge about the initial state. Expectation-Maximization\footnote{An expectation-maximization algorithm can be viewed as performing free-energy minimization \cite{murphey_book,koller_book}. In the E-step, the free-energy is computed and the M-step updates the parameters to minimize the free-energy.} can be used to find the optimal parameters of our model: the policy $\pi$. The E-step will, for a given $\pi$, compute a posterior over state-action sequences. The M-step then adapts the model parameters $\pi$ to optimize the expected likelihood with respect to the quantities calculated in the E-step.

\subsubsection{E-step: a backwards pass in all finite MDPs.}

We use the simpler notation $p(j \mid a, i) \equiv P\left(s_{t+1}=j \mid a_{t}=a, s_{t}=i\right)$ and
$p(j \mid i ; \boldsymbol{\pi})\equiv P\left(s_{t+1}=j \mid s_{t}=i ; \boldsymbol{\pi}\right)=\sum_{a} p(j \mid a, i) \pi_{a i} .$
Further, as a `base' for backward propagation, we define
$$
\begin{aligned}
{\beta}_0(i) &=P\left(c=1 \mid x_{T}=i ; \boldsymbol{\pi}\right) =\sum_a P\left(c=1 \mid a_{T}=a, x_{T}=i\right) \pi_{a i}.
\end{aligned}
$$

This is the immediate cost when following a policy $\pi$. It is the expected cost if there is only one time-step remaining. Then, we can recursively compute all the other backward messages. We use the index $\tau$ to indicate a backwards counter. This means that $\tau + t = T$, where $T$ is total (unknown) horizon length. This is computed as

$$
\beta_{\tau}(i) =P\left(c=1 \mid x_{T-\tau}=i ; \boldsymbol{\pi}\right) =\sum_{j} p(j \mid i ; \boldsymbol{\pi}) \beta_{\tau-1}(j).
$$

Intuitively, the backward messages are the expected cost if one incurs a cost at the last time step only. So, $\beta_2$, is the expected cost if the agent follows the policy $\pi$ for two time-steps and only incurs a cost after that. Using these messages, we can compute a value function dependent on time, actions and states given as: 

$$
\begin{aligned}
q_{\tau}(a, i) &=P\left(c=1 \mid a_{t}=a, s_{t} =i, T=t+\tau ; \boldsymbol{\pi}\right) \\
&=\left\{\begin{array}{ll}
\sum_{j} p(j \mid i, a) \beta_{\tau-1}(j) & \tau>1 \\
P\left(c=1 \mid a_{T}=a, s_{T}=i\right) & \tau=0.
\end{array}\right.
\end{aligned}
$$

\noindent Marginalizing out time, we get the state-action value-function
$$
\begin{aligned}
&P\left(c=1 \mid a_{t}=a, s_{t}=i ; \boldsymbol{\pi}\right) =\frac{1}{C} \sum_{\tau} P(T=t+\tau) q_{\tau}(a, i)
\end{aligned}
$$
where $C$ is a normalization constant. This quantity is the probability of getting a maximum cost given a state and action. It is similar to the $Q(s, a)$ function computed in policy iteration.

\subsubsection{M-step: the policy improvement step.}

The standard M-step in an EM-algorithm maximizes the expected complete log-likelihood with respect to the new parameters $\pi'$. Given that the optimal policy for an MDP is deterministic, a greedy M-step can be used. However, our goal is to minimize the log-likelihood in this case as it refers to a the probability of receiving a maximal cost. This can done as

\begin{equation}
\begin{aligned}
&\pi' =\underset{a}{\operatorname{argmin}}( P\left(c=1 \mid a_{t}=a, s_{t}=i ; \boldsymbol{\pi}\right))
\end{aligned}
\end{equation}

This update converges much faster than in a standard M-step. Here an M-step can be used to to obtain a stochastic policy. However, this is unnecessary since the optimal policy is deterministic. Note that there is an infinite number of stochastic policies but a finite number of deterministic ones. In conclusion, a greedy M-step is faster to converge but still guarantees an optimal policy.

\section{Connections between the two views}

\subsection{Exact relationship between policy iteration and planning as probabilistic inference}
\label{subsec:connections_btween_the_2_views}
The messages $\beta$ computed during backward propagation are exactly equal to the value functions for a single MDP of finite time. The full value function is can therefore be written as the sum of the $\beta$s,
$$V_{\pi}(i) =  \sum_T \beta_T(i)$$ 


\noindent since the prior over time $P(T)$ is a uniform prior. If $P(T)$ is not a uniform distribution, this would result in a mixture rather than a sum. The same applies to the relationship between the Q-value function:

$$Q_{\pi}(a, i) = \sum_T q_T(a, i).$$

Hence, the E-step essentially performs a policy evaluation which yields the classical value function. Given this relation to policy evaluation, the M-step performs an operation exactly equivalent to standard policy improvement. Thus, the EM-algorithm using exact inference is equivalent to Policy Iteration but computes the necessary quantities differently.

One unanswered question is when to stop computing the backward messages. In \cite{toussaint2006probabilistic} messages are computed up to a number $T_{max}$. From this perspective, the planning as inference algorithm presented is equivalent to the so-called \textit{truncated policy iteration} algorithm as opposed to the more common \textit{$\epsilon$-greedy} version. 

In the policy evaluation step, one iterates through the state space to update the value $v_{\pi}(s)$ for every state until a termination criterion is met. An $\epsilon$-greedy criterion means that we stop iterating though the state space once the maximum difference in $V_{\pi}(s)$ for any $s$ is smaller than a positive small number $\epsilon$. In truncated policy iteration, however, we iterate through the state space $T_\text{max}$ times and then perform the policy improvement step. The probabilistic inference algorithm presented in this paper is equivalent to truncated policy iteration if we restrict the maximum number of $\beta$ messages to be computed.

\subsection{World states vs temporal states}
\label{subsec: temporal_vs_world_states}
\begin{figure}[!htb]
    \centering
    \includegraphics[width=\textwidth]{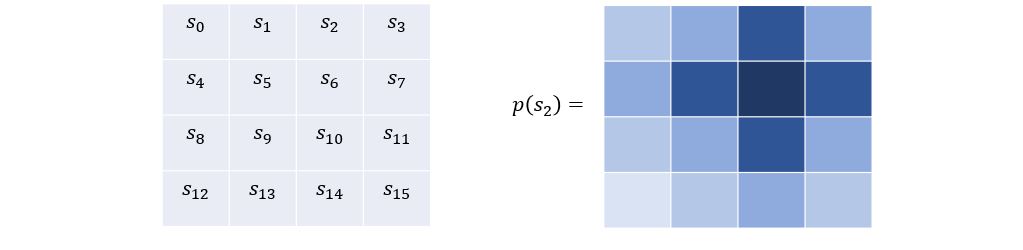}
    \caption{Annotated world states (\textbf{left}) and a posterior over a temporal state (\textbf{right}).}
    \label{fig:tempral_and_world_state}
\end{figure}

In dynamic programming, one reasons over the world states.
In the grid world example in Figure \ref{fig:plan_vs_policy}, this refers to a grid cell. This grid world has 16 world states. In probabilistic inference, one reasons about a\textit{ temporal state}. This is a random variable over all world states. The number of temporal states is dependent on how many time-steps the agents acts in the environment (which is often unknown beforehand). An illustration of the difference is given in Fig. \ref{fig:tempral_and_world_state}.

\subsection{Policies, plans and probabilistic plans}
\label{subsec:plan_policy_prob_plan}

A classical planning algorithm computes a plan: a sequence of actions. An algorithm like A* or Dijkstra's algorithm \cite{cormen2009introduction} can be used to find the optimal path from a stating state to a goal state, given a deterministic world. Crucially, this solution can be computed offline (without interactions with the environment). In a stochastic world, this does not work since the agent can not predict in which states it will end up. However, one can use deterministic planning algorithms for stochastic environments if the path is re-planned online at every time-step.  Determinization-based methods have found success in solving planning under uncertainity problems such as the famous FF-replan algorithm \cite{yoon2007ff-replan}

Active inference approaches computes a \textit{probabilistic plan}. The active inference literature calls this a policy; however, we use a different term to avoid confusion.\footnote{The distinction between a plan and a policy when using active inference has been briefly discussed in \cite{millidge2020relationship}. Additionally, other methods computing plans as probabilistic inference have been proposed before active inference in \cite{attias2003planning,verma2006goal}} 
In active inference, the agents computes a finite plan while interacting with the environment. However, rather than assuming a deterministic world (like FF-replan \cite{yoon2007ff-replan}), the probabilities are taken into account. This can be shown to compute the optimal solution to an MDP (when planning online). We thus refer to it as a probabilistic plan, a plan that was computed while taking the transition probabilities into account.

Finally, a policy is a mapping from states to actions, i.e. the agent has a preferred action to take for every state. Policies can be stochastic or time-dependent; however, for an SSP MDP the optimal policy is deterministic and independent of time. An agent can compute a policy offline and use it online without needing any additional computation while interacting with the environment. The difference between a plan and policy is illustrated in Fig. \ref{fig:plan_vs_policy}. 

To summarize, a plan or a probabilistic plan can only be used for online planning. Since the outcome of an action is inherently uncertain. Probabilistic plans (as used in active inference) find an optimal solution when used to plan online. A policy also provides an optimal solution and can be computed offline or online.



\section{Discussion}
\label{sec:discussion}

In this paper we present a novel approach to solve a stochastic shortest path Markov decision process (SSP MDP) as probabilistic inference. The SSP MDP generalizes many models, including finite and infinite MDPs. Crucially, the dynamic programming algorithms (such as policy iteration) classically used to solve an SSP MDP are valid for \textit{indefinite horizons} (finite but of unknown length); this is not the case for active inference approaches. 

The exact connections between solving an MDP using policy iteration and the presented algorithm are discussed.
Afterwards, we discussed the gap between solving an MDP in active inference and the approaches in the artificial intelligence community. This included the difference between world states and temporal states, the difference between plans, probabilistic plans and policies. An interesting question now is, which approach is more appropriate? This depends on the problem at hand and whether it can be solved online or offline.\\

\noindent \textbf{{Online and offline planning.}
}As discussed in Section \ref{subsec:plan_policy_prob_plan}, a policy is mapping from states to actions and can be used for offline and online planning. Computing a policy is somewhat computationally expensive; however, a look-up is very cheap. Thus if one operates in an environment where the transition and cost function do not change, it is best to compute an optimal policy offline then use it online (while interacting with the environment). This is the case for many planning and scheduling problems, such as a set of elevators operating in sync \cite{crites1996elevator}, task-level planning in robotics \cite{lacerda2019probabilistic}, multi-objective planning \cite{painter2020convex,etessami2007multi-objective} and playing games \cite{Campbell2002DeepB,silver2017mastering}. The challenges in these problems are often that the state-space is incredibly large and thus approximations are needed. However, the problem is fully observable and the cost and transition models are static; the rules of chess do not change half way, for instance.

If the transition or cost functions vary while interacting with the environment (e.g. \cite{corrado_bt,duckworth2021time}), an offline solution is not optimal. In this case, the agent can plan online by re-evaluating a policy or computing probabilistic plans (as done in active inference). Computing the latter is cheaper and requires less memory. This is because a probabilistic plan is a distribution over actions $p(a_t)$ up to a time horizon $T$ while a (finite policy) is a conditional distribution $p(a_t|s_t)$ over all world states in $S$. For any time-step, the posterior over the action is related to the policy such that $p(a_t) = \sum_s p(a_t|s_t)p(s_t)$. 

Consider the work in \cite{tomy2020battery,corrado_bt}. In both cases a robot operates in an environment susceptible to changes. If the environment changes, the agent can easily construct a new model by varying the cost or transition function but needs to recompute a solution. In \cite{tomy2020battery} the authors recompute a policy at every time-step while in \cite{corrado_bt} a probabilistic plans is computed using active inference. Since in both cases the solution is recomputed at every time-step, active inference is preferred since it requires less memory and can be computationally cheaper.  On the other hand, if the environment only changes occasionally, computing a policy might remain preferable. 

To conclude, if the transition and cost functions ($\mathcal{T}$ and $\mathcal{C}$) are static, it is preferable to compute a policy offline. If $\mathcal{T}$ and $\mathcal{C}$ change occasionally, one may still compute an offline policy and recompute a policy only when a change occurs. However, if the environment is dynamic, computing a probabilistic plan (using active inference) is preferable to recomputing a policy at every time-step.


%
%


%
%
%
\bibliographystyle{splncs04}
\bibliography{myBib}
%

\appendix

\section{Appendix: Illustrations}


\begin{figure}[!htb]
    \centering
    \includegraphics[width=\textwidth]{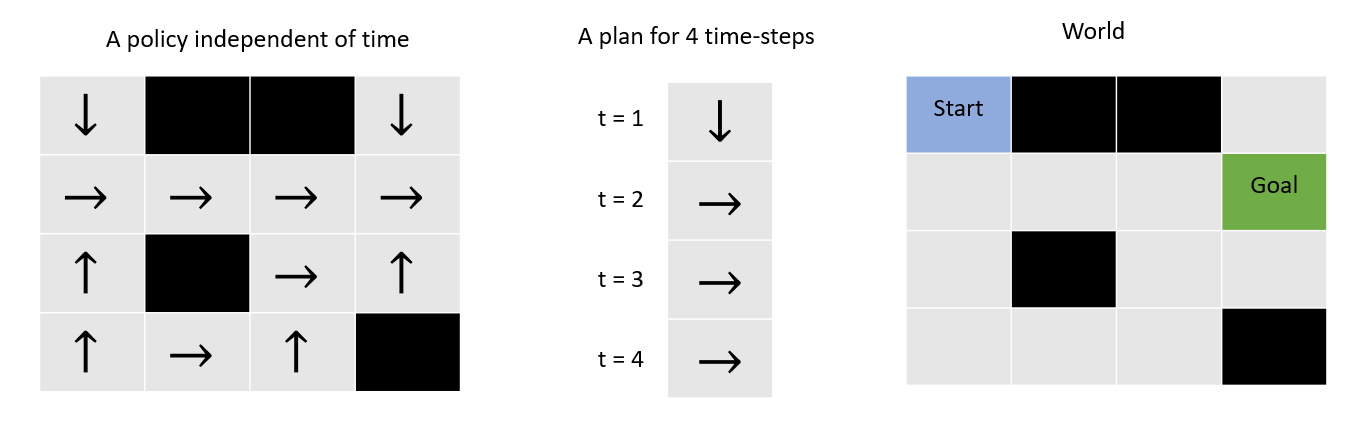}
    \caption{An illustration of a $4\times 4$ grid world \textbf{(right)}. The initial state is blue and goal state is green. An illustration for a policy (\textbf{left}) and a plan (\textbf{middle}). }
    \label{fig:plan_vs_policy}
\end{figure}

\end{document}